\documentclass[11pt]{article}
\PassOptionsToPackage{hyphens}{url}
\usepackage[preprint]{acl}
\usepackage{times}
\usepackage{latexsym}
\usepackage[T1]{fontenc}
\usepackage[utf8]{inputenc}
\usepackage{microtype}
\usepackage{inconsolata}
\usepackage{graphicx}
\usepackage{booktabs}
\usepackage{amsmath}
\usepackage{xspace}
\usepackage{makecell}
\usepackage{multirow}
\usepackage{placeins}
\usepackage{enumitem}
\usepackage{tcolorbox}
\tcbuselibrary{skins}
\definecolor{promptframe}{HTML}{21445B}
\definecolor{promptback}{HTML}{F4F7F9}
\newtcolorbox{promptbox}[1]{enhanced, colback=promptback, colframe=promptframe, coltitle=white, fonttitle=\bfseries\small, title={#1}, boxrule=0.6pt, arc=2pt, left=6pt, right=6pt, top=4pt, bottom=4pt, before upper={\footnotesize\raggedright\setlength{\parskip}{3pt}}}
\newcommand{\pvar}[1]{\textcolor{promptframe}{\textbf{\{#1\}}}}

\usepackage[normalem]{ulem}
\useunder{\uline}{\ul}{}

\newcommand{\name}{\textsc{SoMem}\xspace}
\renewcommand{\dbltopfraction}{0.85}

\title{Can LLM Agents Organize Their Own Memory?}
\author{
  Zichuan Fu\textsuperscript{1}\thanks{Equal contribution.}, Junjia Qi\textsuperscript{1}\footnotemark[1], Wenlin Zhang\textsuperscript{1}, Yimin Deng\textsuperscript{1}, \\
  \textbf{Shirong Wang\textsuperscript{1}, Jinyang Li\textsuperscript{1}, Xian Wu\textsuperscript{2}, Xiangyu Zhao\textsuperscript{1}\thanks{Corresponding author.}} \\
  \textsuperscript{1}City University of Hong Kong, \textsuperscript{2}Tencent Jarvis Lab \\
  \texttt{zc.fu@my.cityu.edu.hk}, \texttt{junjiaqi2-c@my.cityu.edu.hk}, \texttt{wl.z@my.cityu.edu.hk} \\
  \texttt{shirowang6-c@cityu.edu.hk}, \texttt{jinyali2-c@my.cityu.edu.hk} \\
  \texttt{kevinxwu@tencent.com}, \texttt{xianzhao@cityu.edu.hk}
}

\begin{document}
\maketitle

\begin{abstract}

In most LLM agents, weights are fixed after deployment, so new information must be retained in external memory.
Existing approaches often rely on predefined records and fixed organizational rules, which can not readily accommodate evolving tasks and interactions.
This raises the question: can LLM agents organize their own memory using ordinary documents?
We introduce \textbf{s}elf-\textbf{o}rganized \textbf{mem}ory (\name), a framework that stores memory in linked documents.
The agent decides what to summarize, how to group related information, and when to revise existing topics, and every summary it produces stays linked to the original interactions as evidence.
The runtime validates updates and automatically derives temporal, relational, and recent-memory views from the organized content.
At query time, the agent chooses what to search and read, following document links and inspecting original evidence as needed.
On LoCoMo, BEAM-100K, and LongMemEval-S, \name outperforms the strongest baseline by 2.66, 3.10, and 2.00 percentage points, respectively.
The constructed memory can be reused across models, and additional reasoning effort can further improve answer quality.
These findings demonstrate the feasibility of agents organizing and using their own persistent memory through ordinary documents.
Code is available at: https://github.com/Qi202/SoMem.
\end{abstract}

\begin{figure*}[t]
\centering
\includegraphics[width=\textwidth]{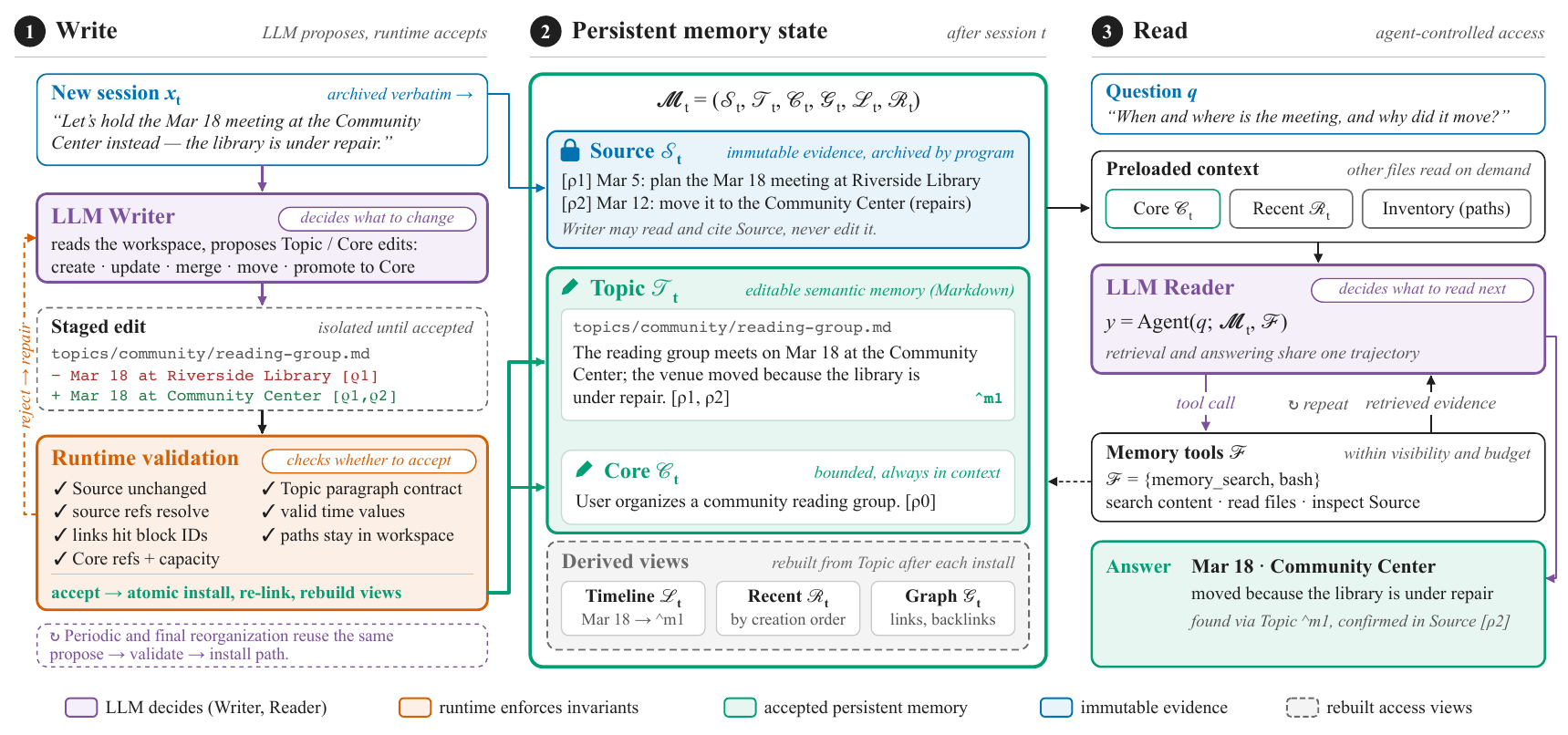}
\caption{Overview of \name, illustrated with a meeting-venue update. The agent revises Topic documents from Source records, and the runtime derives access views from the revised documents. At query time, the agent searches and reads the organized memory and its Source evidence to answer the question. The tools shown are the combined search and read-only shell interface; the main experiments use the equivalent split interface (Appendix~\ref{app:memory_operations}).}
\label{fig:framework}
\end{figure*}

\section{Introduction}
\label{sec:intro}

During deployment, LLM agents encounter new information, including user preferences, task progress, and changes in the environment. This information is not incorporated into their fixed model parameters.
External memory preserves information across interactions and supplies relevant records as context for later requests~\citep{zhong2024memorybank,packer2024memgpt}.
This allows an agent to personalize its responses, continue earlier activities, and use background knowledge acquired through experience~\citep{zhong2024memorybank,park2023generative}.
As conversations extend across sessions, such context can include facts and events that are absent from the current request but necessary for interpreting it~\citep{locomo2024,longmemeval2025}.
This persistent context makes information from earlier interactions available alongside the model's pretrained knowledge.

Designing agent memory requires deciding what to save, how to organize and update it, and how to access it for a task~\citep{memory_survey2025}.
These decisions are connected: the information retained during construction determines the evidence available during retrieval and reading~\citep{longmemeval2025}.
Complete histories preserve detail but require the agent to locate relevant exchanges; compact records reduce the amount of text to inspect but may omit qualifications or context needed later~\citep{longmemeval2025,lightmem2026}.
New interactions can also revise an earlier fact, connect separate events, or change the period during which a statement holds~\citep{longmemeval2025,timem2026}.
A memory system must therefore support both concise access to accumulated information and recovery of the evidence needed to interpret changes.

Existing methods address these requirements through different representations and operations.
MemoryBank and Mem0 extract and update compact records, while Mem0 also offers an entity--relation graph~\citep{zhong2024memorybank,mem0_2024}.
A-Mem generates attributed notes and revises their associations as new memories arrive; LightMem and TiMem consolidate information through processing stages or temporal hierarchies~\citep{amem2025,lightmem2026,timem2026}.
Model-controlled memory operations are also established in MemGPT and the vector-backed semantic file interface of LSFS~\citep{packer2024memgpt,mei2025lsfs}.
Building on such model-controlled operations, we study how editable semantic organization, preserved original evidence, and automatically derived access views can  maintain consistency as memory changes.
The agent revises the semantic content, while the runtime validates evidence references and reconstructs the access views from accepted edits.
The central question is therefore how an agent can revise its memory organization while preserving evidence and consistent access.

We propose \textbf{s}elf-\textbf{o}rganized \textbf{mem}ory (\name), in which the model controls semantic organization and the runtime maintains evidence and access consistency.
First, the agent chooses how to group and revise information in Topic documents, which reference immutable Source records and from which a small Core is promoted for persistent context.
Second, the runtime applies the same validation checks to every revision. Once a revision is accepted, it rebuilds the access views from the updated content and metadata.
This allows semantic grouping to change without requiring the model to maintain separate versions for each access structure.
Third, the agent selects further searches and reads from its observations, returning to original records when organized content is insufficient.
The same agent can write and read memory, and other models can reuse the resulting workspace through the same interface.

Our contributions are as follows:
\begin{itemize}[leftmargin=*, itemsep=0pt]
    \item We formulate self-organized memory as editable semantic content that retains links to its supporting evidence. Let the model decide how that content is grouped, while the runtime maintains the access structures that serve retrieval.
    \item We develop a reorganization procedure that validates edits and reconstructs access views, coupled with sequential inspection of the evidence needed to answer a question.
    \item We demonstrate the effectiveness of \name through comparisons with baselines on LoCoMo, BEAM-100K, and LongMemEval-S, using a shared GPT-6 Luna Reader. We also show that \name works across multiple backbones and that the constructed memory can be reused across models.
\end{itemize}

\section{Method}
\label{sec:method}
\label{subsec:overview}

\name lets an agent organize persistent memory through editable Markdown documents (Figure~\ref{fig:framework}).
The agent uses Source records to construct and revise Topic documents, from which a small Core is promoted for persistent context.
Deterministic code derives Graph, Timeline, and Recent views from Topic. The agent retrieves evidence using search and document access tools. Implementation details are provided in the Appendix~\ref{app:memory_operations}.

\subsection{Self-Organizing Memory Documents}
\label{subsec:workspace}

Information about the same person, activity, or task can accumulate across many interactions, and later observations may change how earlier information should be grouped.
Predefined memory schemas constrain how this grouping can change~\citep{amem2025}, while storing information as isolated records can leave related evidence scattered across separate items~\citep{infinimemory2026}.
We therefore let the agent organize memory in editable, linked documents whose subjects and internal structure can evolve with the content.

At update $t$, the system archives new interaction records $x_t$ as Source records, preserving their original content for later reference.
The agent uses Source and existing memory to update Topic and Core:
\begin{equation}
\begin{aligned}
\mathcal S_t &= \operatorname{Append}(\mathcal S_{t-1},x_t),\\
(\mathcal T_t,\mathcal C_t)
&=\operatorname{Organize}(\mathcal S_t,\mathcal T_{t-1},\mathcal C_{t-1}).
\end{aligned}
\label{eq:memory-update}
\end{equation}
Here, $\mathcal S_t$ contains the archived Source records, and $\mathcal T_t$ is the Topic document collection. Core, denoted by $\mathcal C_t\in\mathcal T_t$, is a size-bounded Markdown file within this collection. The agent promotes long-term preferences, standing requirements, and frequently needed background from Topic documents into Core. Core is placed in the initial context of every query.
The agent reads relevant Source records and existing documents to decide what to retain and where it belongs.
It can create or revise paragraphs and documents, adjust headings, split broad topics, and regroup related information.
Edits are staged before acceptance.
The runtime checks Source immutability, document structure, and the resolution of evidence references and links, then installs accepted edits and rebuilds the derived views; rejected edits preserve the prior memory state.
Periodic organization revisits changed Topic files, while an optional final pass reorganizes the complete Topic collection.

\begin{figure}[t]
\centering
\includegraphics[width=\columnwidth]{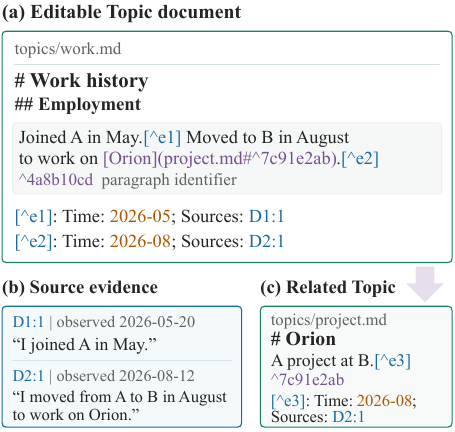}
\caption{An example Topic document. Each statement is annotated with its event time and supporting Source records, and paragraph references link related Topic documents across the workspace.}
\label{fig:document_format}
\end{figure}

Figure~\ref{fig:document_format} illustrates this structure with two employment events in a single Topic paragraph. Each event has its own time annotation and Source references. A link within the paragraph points to a specific paragraph in the related project document.
The agent determines each Markdown document's topic and internal structure.
Within each document, it uses headings to define subtopics and groups related information into paragraphs.
It also links specific paragraphs across documents through references.
A paragraph is a memory unit; unit $i$ and its evidence annotations are represented as:
\begin{equation}
\begin{aligned}
u_i &= (b_i,c_i,E_i,L_i),\\
E_i &= \{(s_{ij},\tau_{ij},R_{ij})\}_{j=1}^{n_i}.
\end{aligned}
\label{eq:document-unit}
\end{equation}
Here, $b_i$ is a stable paragraph identifier, $c_i$ is the paragraph text, and $L_i$ contains the identifiers of referenced memory units.
Each of the $n_i$ annotations associates a statement span $s_{ij}$ in $c_i$ with an event time $\tau_{ij}$ and supporting references $R_{ij}$ to Source records in $\mathcal S_t$.
In Markdown, a footnote marker follows the corresponding statement; its definition records the time and source links, and a trailing paragraph identifier makes the unit addressable from other documents.
Event times preserve the temporal precision of the supporting Source records: year, month, or day.

\subsection{Automatic Memory View Derivation}
\label{subsec:organization}

Questions may access memory by relation, time, or recency~\citep{longmemeval2025}, which a single topical organization does not directly support.
Maintaining a separate representation for each access pattern, however, requires propagating every revision across them and increases update overhead~\citep{lightmem2026}.
In \name, the runtime derives access views from the Topic documents maintained by the agent, using paragraph references, event times, and the creation order of memory units:
\begin{equation}
\begin{aligned}
\mathcal G_t &= \operatorname{Graph}(\mathcal T_t),\\
\mathcal L_t &= \operatorname{Timeline}(\mathcal T_t),\\
\mathcal R_t &= \operatorname{Recent}_{K}(\mathcal T_t).
\end{aligned}
\label{eq:derived-views}
\end{equation}
Here, $\mathcal G_t$, $\mathcal L_t$, and $\mathcal R_t$ denote Graph, Timeline, and Recent, respectively.
These operations construct the views without requiring the model to write each representation separately.

\paragraph{Graph.}
Each Topic memory unit is a node identified by $b_i$; a reference $b_j\in L_i$ defines a directed edge from unit $i$ to unit $j$.
The system parses these references into a Relations index that records outgoing links and backlinks for each node.
The agent can therefore access both the memories cited by a paragraph and those that cite it.
Graph thus uses the existing Topic structure without a separate relation extraction step.

\paragraph{Timeline.}
For each evidence annotation, the system groups its statement $s_{ij}$ into a Markdown file according to the event time $\tau_{ij}$, retaining its Source references $R_{ij}$.
File paths preserve the original temporal precision: years, months, and days correspond to \nolinkurl{timeline/YYYY.md}, \nolinkurl{timeline/YYYY/MM.md}, and \nolinkurl{timeline/YYYY/MM/DD.md}, respectively.
Each file begins with a level-one heading specifying the corresponding time, followed by individual event entries.
Each entry contains the event statement, a reference to its original Topic paragraph, and references to the supporting Source messages.

\paragraph{Recent.}
The system records the creation order of each memory unit.
$\operatorname{Recent}_{K}$ uses this order to select the $K$ most recently created memory units from Topic. The resulting view retains each unit's paragraph content, Topic location, and source references.
Recent updates automatically with memory writes and is included in the initial context of each query.

\subsection{Multi-View Memory Retrieval}
\label{subsec:query}

Relevant evidence depends on the question: a topic match may miss temporal constraints, while a compact fact may omit needed context~\citep{longmemeval2025}.
Such evidence is often identified only after related memory has been read, so a single retrieval before answering may be insufficient.
\name provides read-only operations for content search, file listing, and document reading:
\[
\mathcal F=\{\operatorname{Search},\operatorname{List},\operatorname{Read}\}.
\]

\paragraph{Content search.}
The agent retrieves Topic and Source passages with BM25~\citep{robertson2009bm25} lexical matching or dense semantic search, choosing the mode for each query.
Optional path and date-range filters restrict the search to relevant documents or periods.
Each result provides a passage and its location, allowing the agent to inspect the surrounding context.

\paragraph{Document and view access.}
File-listing and reading operations let the agent locate documents and inspect their contents, including selected line ranges.
The agent can inspect Topic sections, read Timeline files for relevant dates, or consult Graph links and backlinks to locate related paragraphs.
Source references in Topic and Timeline allow it to check statements against the original interactions.
These operations can be combined according to the question, without a prescribed retrieval sequence.

Given a question $q$, the agent receives Core, Recent, and a file inventory as its initial memory context. The question date is also provided when available.
It uses these tools to retrieve relevant memory and generate an answer supported by the evidence:
\begin{equation}
\begin{aligned}
\mathcal M_t&=(\mathcal S_t,\mathcal T_t,\mathcal C_t,
\mathcal G_t,\mathcal L_t,\mathcal R_t),\\
y&=\operatorname{Agent}(q;\mathcal M_t,\mathcal F).
\end{aligned}
\label{eq:query}
\end{equation}
Here, $\mathcal M_t$ comprises the documents and views constructed above, $\mathcal F$ denotes the available read-only access operations, and $y$ is the generated answer.
The documents and views remain available for further inspection through the same tool interface.

\section{Experiments}
\label{sec:exp}

We compare memory methods in their full configurations on LoCoMo, BEAM-100K, and LongMemEval-S, then examine component ablations, cross-model memory reuse, and Reader effort.

\subsection{Experimental Setup}

\paragraph{Datasets.}
We evaluate \name on three long-term memory benchmarks: LoCoMo~\citep{locomo2024}, LongMemEval-S~\citep{longmemeval2025}, and BEAM-100K~\citep{beam2025}.
LoCoMo tests single-hop (SH), multi-hop (MH), open-domain (OD), and temporal (TR) reasoning over multi-session histories; LongMemEval-S covers knowledge update (KU), multi-session reasoning (MS), single-session assistant (SSA), preference (SSP), and user (SSU) information, and temporal reasoning (TR); BEAM-100K covers ten memory capabilities: information extraction (IE), multi-session reasoning (MR), temporal reasoning (TR), knowledge update (KU), contradiction resolution (CR), event ordering (EO), instruction following (IF), preference following (PF), summarization (SUM), and abstention (ABS).
Table~\ref{tab:dataset_scope} summarizes the evaluation and analysis-subset sizes.
The main method comparisons use the complete evaluation sets; component ablations use the full LoCoMo set, and Reader-effort analyses use fixed subsets.
Subsets are stratified by conversation and question type and selected independently of model outputs and scores. Within each analysis, all conditions use the same subsets.

\begin{table}[t]
\centering
\small
\setlength{\tabcolsep}{4pt}
\caption{Statistics of the evaluation datasets, including full evaluation sets and Reader-effort subsets. Component ablations use the full LoCoMo set.}
\label{tab:dataset_scope}
\begin{tabular*}{\columnwidth}{@{\extracolsep{\fill}}llr@{}}
\toprule
\textbf{Dataset} & \textbf{Task scope} & \textbf{Full / subset} \\
\midrule
LoCoMo & 4 QA types & 1,540 / 100 \\
LongMemEval-S & 6 QA types & 500 / -- \\
BEAM-100K & 10 capabilities & 400 / 200 \\
\bottomrule
\end{tabular*}
\end{table}

\paragraph{Evaluation Metrics.}
We use GPT-4o-mini as the judge with temperature~0 to evaluate complete, unmodified answers.
For LoCoMo and LongMemEval-S, the judge assesses correctness based on the question, reference answer, and task-specific requirements. We report LLM-judge (LJ) accuracy by question type and overall.
Overall accuracy is the proportion of correctly answered questions across the evaluation set.
For BEAM-100K, each rubric criterion receives a score of 0, 0.5, or 1; we average criteria within each question and then average over questions to obtain the overall nugget score.
We also report the pass rate, counting questions with a score of at least 0.5 as passes.
Evaluation protocols and configurations are provided in Appendix~\ref{app:protocol}.

\paragraph{Implementation Details.}
We compare \name with representative memory methods: A-Mem~\citep{amem2025}, Mem0~\citep{mem0_2024}, MemoryOS~\citep{memoryos2025}, Nemori~\citep{nemori2026}, LightMem~\citep{lightmem2026}, and E-mem~\citep{emem2026}.
In the primary comparisons, GPT-6 Luna serves as both the memory-construction model (Writer) and the question-answering model (Reader) for every memory method, while each method retains its own memory organization and retrieval pipeline; no Reader uses reasoning effort.
Full Context uses the same Reader and answers directly from the complete original history, without truncation or retrieval tools.
We additionally report \name with other backbones, each serving as both Writer and Reader; these rows are excluded from the ranking.
Bold and underline mark the best and second-best distinct values among methods with the shared Reader, including ties.
The cross-model and Reader-effort analyses further use GPT-5.6 Luna, Gemini 3 Flash, DeepSeek V4 Flash, Qwen 3.8 Flash, GPT-6.1 Sol, and GLM-5.3 Flash.
Additional implementation details and evaluation protocols are provided in Appendices~\ref{app:memory_operations} and~\ref{app:protocol}.

\subsection{Main Results}
\label{sec:exp:main}

\begin{table}[t]
\centering
\small
\setlength{\tabcolsep}{3pt}
\caption{LoCoMo LJ accuracy (\%) on all 1,540 category 1--4 questions.}
\label{tab:main_results}
\begin{tabular*}{\columnwidth}{@{\extracolsep{\fill}}lccccc@{}}
\toprule
\textbf{Method} & \textbf{SH} & \textbf{MH} & \textbf{OD} & \textbf{TR} & \textbf{Overall} \\
\midrule
Full Context & \underline{94.17} & 80.85 & 45.83 & 72.59 & 84.22 \\
\midrule
A-Mem & 62.78 & 55.67 & 41.67 & 62.31 & 60.06 \\
Mem0 & 83.83 & 76.95 & 53.12 & \underline{82.24} & 80.32 \\
MemoryOS & 60.05 & 56.03 & 43.75 & 50.16 & 56.23 \\
Nemori & 79.79 & 67.73 & 43.75 & 66.36 & 72.53 \\
LightMem & 82.88 & 70.21 & 52.08 & 79.75 & 77.99 \\
E-mem & 92.75 & \textbf{85.11} & \underline{58.33} & 78.82 & \underline{86.30} \\
\textbf{\name} & \textbf{94.53} & \underline{83.33} & \textbf{59.38} & \textbf{88.16} & \textbf{88.96} \\
\textsc{SoMem}$^{\dagger}$ & 95.96 & 93.26 & 65.62 & 91.90 & 92.73 \\
\textsc{SoMem}$^{\ddagger}$ & 95.36 & 91.49 & 72.92 & 90.97 & 92.34 \\
\textsc{SoMem}$^{\S}$ & 94.89 & 91.13 & 63.54 & 89.72 & 91.17 \\
\bottomrule
\end{tabular*}
\par\smallskip\parbox{\columnwidth}{\footnotesize Writer and Reader: $^{\dagger}$\,DeepSeek V4 Flash; $^{\ddagger}$\,GPT-5.6 Luna; $^{\S}$\,Gemini 3 Flash. These rows are excluded from the ranking; all other rows use GPT-6 Luna.}
\end{table}

Across the benchmarks (Tables~\ref{tab:main_results}--\ref{tab:longmemeval}), \name achieves the highest overall scores among the methods evaluated with the shared GPT-6 Luna Reader. The comparisons include direct full-context answering.
In Table~\ref{tab:main_results}, its advantages in single-hop recall and temporal reasoning highlight the effectiveness of locating specific facts and retaining their context.
Topic organization makes related information accessible, while links to Source allow the agent to retrieve the precise details needed to answer these questions.
Results with additional backbones also demonstrate the applicability of this representation across models.

\begin{table*}[t]
\centering
\small
\setlength{\tabcolsep}{4pt}
\caption{BEAM-100K results (\%) on 400 questions. Capability columns report mean nugget scores; the final columns report overall nugget score and pass rate.}
\label{tab:beam_method_comparison}
\begin{tabular*}{\textwidth}{@{\extracolsep{\fill}}lcccccccccccc@{}}
\toprule
\textbf{Method} & \textbf{IE} & \textbf{MR} & \textbf{TR} & \textbf{KU} & \textbf{CR} & \textbf{EO} & \textbf{IF} & \textbf{PF} & \textbf{SUM} & \textbf{ABS} & \textbf{Nugget} & \textbf{Pass} \\
\midrule
Full Context & \textbf{82.86} & 50.25 & \underline{41.25} & \underline{50.00} & 25.63 & 36.95 & 56.25 & 84.38 & \textbf{49.14} & \underline{32.50} & \underline{50.92} & 55.25 \\
\midrule
A-Mem & \underline{76.82} & \underline{54.30} & \underline{41.25} & 46.25 & 23.44 & 39.78 & 66.88 & \underline{86.25} & 40.11 & 30.00 & 50.51 & \underline{55.75} \\
Mem0 & 62.29 & 50.50 & 28.75 & \textbf{58.75} & 30.00 & \underline{43.99} & \underline{68.12} & \textbf{86.88} & 35.70 & 25.00 & 49.00 & 53.25 \\
MemoryOS & 60.94 & \textbf{60.49} & 34.38 & 39.38 & 26.88 & 32.93 & 63.75 & 82.08 & 36.24 & \textbf{42.50} & 47.96 & 50.75 \\
Nemori & 63.96 & 48.59 & 24.38 & 46.25 & 24.06 & 41.41 & \textbf{73.75} & 85.00 & 44.72 & 27.50 & 47.96 & 51.50 \\
LightMem & 62.29 & 52.88 & 31.87 & 42.50 & 30.31 & 29.72 & 61.25 & 77.50 & 18.37 & 28.75 & 43.54 & 47.00 \\
E-mem & 59.58 & 45.36 & 30.00 & 31.25 & \textbf{51.56} & 41.96 & 54.37 & 76.25 & 25.77 & 17.50 & 43.36 & 49.00 \\
\textbf{\name} & 73.70 & 51.53 & \textbf{53.75} & 48.75 & \underline{36.56} & \textbf{46.84} & \textbf{73.75} & 85.62 & \underline{45.94} & 23.75 & \textbf{54.02} & \textbf{61.50} \\
\bottomrule
\end{tabular*}
\end{table*}

Table~\ref{tab:beam_method_comparison} shows strengths in temporal reasoning, event ordering, and instruction following over long histories.
These tasks require connecting events or requirements scattered across interactions and interpreting them in context.
Organizing related content by topic and linking it to time-stamped original evidence supports this process: relevant statements can be located together and checked against the history in which they occurred.
This provides a coherent basis for relating distant events and recovering instructions whose meaning depends on earlier exchanges.

\begin{table*}[t]
\centering
\small
\setlength{\tabcolsep}{4pt}
\caption{LongMemEval-S LJ accuracy (\%) on all 500 questions.}
\label{tab:longmemeval}
\begin{tabular*}{\textwidth}{@{\extracolsep{\fill}}lccccccc@{}}
\toprule
\textbf{Method} & \textbf{KU} & \textbf{MS} & \textbf{SSA} & \textbf{SSP} & \textbf{SSU} & \textbf{TR} & \textbf{Overall} \\
\midrule
Full Context & \textbf{92.31} & 51.13 & \underline{96.43} & 53.33 & 90.00 & 42.11 & 65.80 \\
\midrule
A-Mem & 75.64 & 49.62 & \underline{96.43} & 30.00 & \underline{92.86} & 45.86 & 62.80 \\
Mem0 & 91.03 & \underline{60.90} & 76.79 & \underline{73.33} & 91.43 & 62.41 & 72.80 \\
MemoryOS & 73.08 & 36.09 & 87.50 & 66.67 & 72.86 & 30.08 & 53.00 \\
Nemori & 79.49 & 51.88 & 82.14 & 33.33 & \underline{92.86} & 60.15 & 66.40 \\
LightMem & \underline{91.67} & \textbf{75.56} & 25.00 & \textbf{86.67} & \underline{92.86} & \underline{71.80} & \underline{74.50} \\
E-mem & 87.18 & 46.62 & 94.64 & 26.67 & 85.71 & 56.39 & 65.20 \\
\textbf{\name} & 77.56 & 58.65 & \textbf{99.11} & \underline{73.33} & \textbf{98.57} & \textbf{73.31} & \textbf{76.50} \\
\textsc{SoMem}$^{\dagger}$ & 91.03 & 83.46 & 100.00 & 76.67 & 100.00 & 90.23 & 90.20 \\
\bottomrule
\end{tabular*}
\par\smallskip\parbox{\textwidth}{\footnotesize Writer and Reader: $^{\dagger}$\,DeepSeek V4 Flash. This row is excluded from the ranking; all other rows use GPT-6 Luna.}
\end{table*}

In Table~\ref{tab:longmemeval}, \name is particularly effective at recalling specific information from both user and assistant interactions.
Preserving both sides of the original dialogue keeps details omitted from topic summaries available for later inspection and helps answer questions that require precise historical content.
The observed strengths highlight the value of retaining detailed evidence alongside organized summaries for long-term conversational recall.

\subsection{Ablation Study}
\label{sec:exp:ablation}

Table~\ref{tab:ablation_locomo} compares Full with component removals on all LoCoMo evaluation questions.
We freeze memory and fix the Reader (GPT-6 Luna without reasoning effort) and query settings, changing only query-time access; w/o Topic restricts the Reader to Source records.
Additional settings, context measurements, and paired results are provided in Appendix~\ref{app:full_ablation_results}.

\begin{table}[t]
\centering
\small
\setlength{\tabcolsep}{4pt}
\caption{LoCoMo component ablations on all 1,540 questions (LJ accuracy, \%). Bold marks the Full reference, not the best score.}
\label{tab:ablation_locomo}
\begin{tabular*}{\columnwidth}{@{\extracolsep{\fill}}lccccc@{}}
\toprule
\textbf{Condition} & \textbf{SH} & \textbf{MH} & \textbf{OD} & \textbf{TR} & \textbf{Overall} \\
\midrule
\textbf{Full} & \textbf{94.53} & \textbf{83.33} & \textbf{59.38} & \textbf{88.16} & \textbf{88.96} \\
\midrule
w/o Topic & 92.51 & 76.95 & 57.29 & 85.67 & 86.04 \\
w/o Core & 93.10 & 86.52 & 58.33 & 89.41 & 88.96 \\
w/o Source & 83.35 & 82.98 & 52.08 & 78.50 & 80.32 \\
w/o Recent & 92.51 & 81.91 & 57.29 & 89.10 & 87.66 \\
w/o Timeline & 93.22 & 85.46 & 53.12 & 86.92 & 87.99 \\
w/o Graph & 92.87 & 84.04 & 61.46 & 86.92 & 88.05 \\
\bottomrule
\end{tabular*}
\end{table}

Removing either Topic or Source reduces overall accuracy, with paired confidence intervals excluding zero in both cases (Table~\ref{tab:ablation_paired}), supporting the complementary roles of organized content and original evidence.
Topic groups dispersed facts into thematic entry points, and Source preserves the original details and context needed to verify them.
Removing Source produces the larger decline, indicating that organized content complements rather than replaces the original records.

The remaining components mainly shift accuracy across question types: removing Core or any single derived view changes overall accuracy by at most 1.30 points, and the paired intervals include zero (Table~\ref{tab:ablation_paired}).
Core is a small, bounded subset of Topic that holds long-term preferences, standing requirements, and frequently needed background, and that the Reader receives at the start of every query, much like a system prompt. LoCoMo contains no question type dedicated to preferences or standing instructions, so removing Core leaves overall accuracy unchanged; its role is to carry persistent constraints rather than to supply factual evidence.
Removing Recent lowers single-hop accuracy from 94.53 to 92.51 while raising temporal accuracy from 88.16 to 89.10.
Recent preloads entries selected by memory creation order, providing readily available factual cues for direct recall.
Removing this preload substantially shortens the input to the final answer-producing call while leaving the average call count similar (Table~\ref{tab:ablation_resources}).
The Reader can therefore continue to inspect Topic and Source with a more compact input, consistent with the different evidence-selection demands of temporal questions.
Removing Timeline or Graph each lowers temporal accuracy to 86.92, consistent with time-ordered entries and cross-document links both helping to relate events.
Together, these results indicate that Topic and Source are the main contributors to overall accuracy. The derived views provide access paths for specific question types, while Core carries persistent constraints that LoCoMo rarely tests. The effects of the derived views and Core on overall accuracy remain within sampling uncertainty.

\subsection{Impact of Backbone Models}
\label{sec:exp:cross_model}

We evaluate Writer--Reader combinations on LoCoMo, using GPT-5.6 Luna, Gemini 3 Flash, and DeepSeek V4 Flash as Writers and these models together with Qwen 3.8 Flash as Readers. We compare memories built by different Writers with a fixed Reader and different Readers on the same memory.

\begin{table*}[t]
\centering
\small
\setlength{\tabcolsep}{4pt}
\caption{LoCoMo accuracy (\%) for Writer--Reader combinations on all 1,540 questions. Each Writer's frozen memory is reused by all Readers. Bold marks the best Reader for each Writer and task, including ties.}
\label{tab:writer_reader_matrix}
\begin{tabular*}{\textwidth}{@{\extracolsep{\fill}}llccccc@{}}
\toprule
\textbf{Writer} & \textbf{Reader} & \textbf{SH} & \textbf{MH} & \textbf{OD} & \textbf{TR} & \textbf{Overall} \\
\midrule
\multirow{4}{*}{\shortstack[l]{GPT-5.6\\Luna}} & GPT-5.6 Luna & 95.36 & 91.49 & 72.92 & 90.97 & 92.34 \\
 & Gemini 3 Flash & 96.67 & 91.49 & 67.71 & 91.28 & 92.79 \\
 & DeepSeek V4 Flash & \textbf{96.79} & 92.91 & 69.79 & 90.97 & 93.18 \\
 & Qwen 3.8 Flash & 95.01 & \textbf{93.97} & \textbf{73.96} & \textbf{93.77} & \textbf{93.25} \\
\midrule
\multirow{4}{*}{\shortstack[l]{Gemini 3\\Flash}} & GPT-5.6 Luna & 96.08 & 91.13 & 64.58 & 90.03 & 91.95 \\
 & Gemini 3 Flash & 94.89 & 91.13 & 63.54 & 89.72 & 91.17 \\
 & DeepSeek V4 Flash & \textbf{96.79} & \textbf{94.68} & 60.42 & \textbf{91.90} & \textbf{93.12} \\
 & Qwen 3.8 Flash & 94.65 & 90.07 & \textbf{66.67} & 90.65 & 91.23 \\
\midrule
\multirow{4}{*}{\shortstack[l]{DeepSeek V4\\Flash}} & GPT-5.6 Luna & 95.24 & 91.49 & 69.79 & 90.65 & 92.01 \\
 & Gemini 3 Flash & \textbf{96.91} & 91.84 & 68.75 & 90.34 & \textbf{92.86} \\
 & DeepSeek V4 Flash & 95.96 & \textbf{93.26} & 65.62 & \textbf{91.90} & 92.73 \\
 & Qwen 3.8 Flash & 92.39 & 89.01 & \textbf{70.83} & 90.97 & 90.13 \\
\bottomrule
\end{tabular*}
\end{table*}

As shown in Table~\ref{tab:writer_reader_matrix}, memory built by GPT-5.6 Luna achieves higher overall accuracy with most Readers, and its accuracy averaged across Readers is highest for every task category.
The usefulness of the constructed memory therefore extends beyond the model that produced it: other Readers can use the same documents and evidence references without rebuilding memory.

Reader choice also changes the task profile of the same memory.
DeepSeek V4 Flash achieves higher multi-hop accuracy than GPT-5.6 Luna and Gemini 3 Flash across the evaluated memories, whereas Qwen 3.8 Flash consistently achieves the highest open-domain accuracy.
Multi-hop questions require combining recorded facts, while open-domain questions require inferences from conversational context; these distinct requirements make the task breakdown more informative than a single overall Reader ranking.
Qwen 3.8 Flash's lower single-hop accuracy also explains why its open-domain advantage does not consistently yield the highest overall score: single-hop questions account for a much larger share of LoCoMo.

These results support choosing Writers and Readers separately according to task requirements.
A constructed workspace can serve Readers with different task profiles, allowing the answering model to change while the stored memory is reused.

\subsection{Scaling}
\label{sec:exp:scaling}

We vary the Reader's requested reasoning effort while reusing constructed memory and fixed question sets within each sweep.
We measure actual usage as the mean cumulative input and output tokens across Reader calls per question, excluding memory construction and evaluation.

\begin{figure*}[t]
\centering
\includegraphics[trim=0 0 0 20,clip,width=\textwidth]{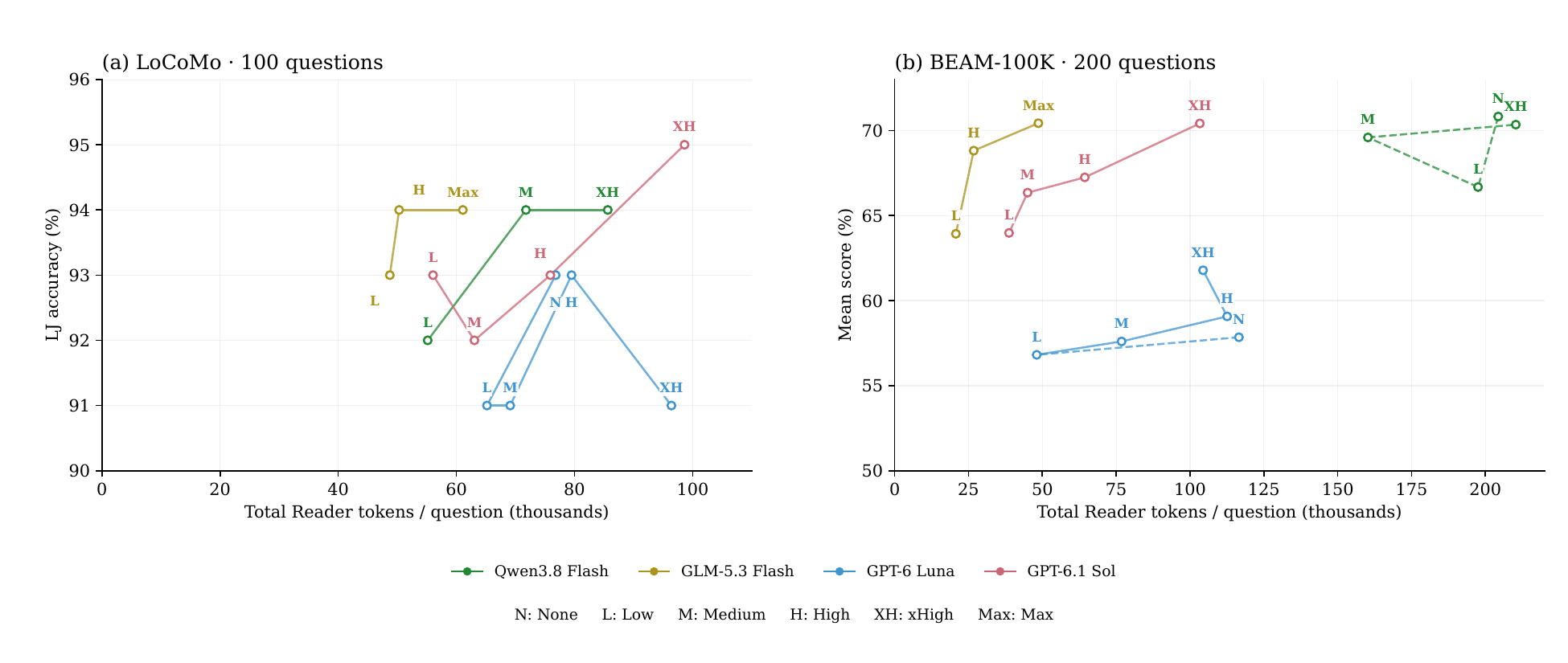}
\caption{Reader scaling on fixed LoCoMo (100 questions) and BEAM-100K (200) subsets. The x-axis is mean cumulative Reader input and output tokens per question, including cached and repeated input. LoCoMo reports LJ accuracy; BEAM reports mean nugget score ($\times100$). Colors denote models and point labels denote requested effort.}
\label{fig:reader_compute}
\end{figure*}

Increasing reasoning effort generally improves answer quality, with more sustained gains on BEAM's longer histories (Figure~\ref{fig:reader_compute}).
These tasks require locating clues, following up on original records, and integrating evidence distributed across the history, leaving more scope for additional reasoning to support evidence acquisition and synthesis.
By allowing the Reader to continue searching and reading as needed, \name provides a way to use additional inference resources over existing memory.

The gains nevertheless exhibit saturation and occasional reversals.
On LoCoMo, some Readers continue to improve at higher effort levels, whereas others plateau or decline at the highest tested setting.
Additional computation therefore does not necessarily translate into more effective evidence use: once the available information is sufficient, further reasoning may offer limited benefit.
The appropriate budget depends on both the Reader and the task, rather than always favoring the highest effort level.

Higher reasoning effort need not increase total token usage either.
On BEAM-100K, increasing GPT-6 Luna's requested reasoning effort from High to xHigh improves answer quality while reducing searches, document reads, Reader calls, and cumulative token usage.
This pattern is consistent with more targeted query and reading decisions reducing repeated lookups and the repeated input of accumulated context across calls.
Thus, stronger reasoning may not only improve answer quality but also reduce token consumption through more efficient memory access, potentially lowering usage costs.

\section{Related Work}
\label{sec:related}

\subsection{Memory Representation and Organization}

Long-term memory systems differ in what they store from an interaction and how they connect it to earlier information.
MemoryBank~\citep{zhong2024memorybank} and Mem0~\citep{mem0_2024} retain compact records, and LangMem~\citep{langmem2025} provides tools with which an agent adds, updates, deletes, and searches memories, complemented by background extraction and consolidation.
Nemori~\citep{nemori2026} decides what to retain by how poorly an experience is predicted, and A-Mem~\citep{amem2025} adds attributes and associative links to notes.
MemMachine~\citep{memmachine2026} instead stores raw conversational episodes with limited LLM-based extraction and retrieves them together with their surrounding dialogue.
LightMem~\citep{lightmem2026} separates memory-processing stages to reduce online cost.
TiMem~\citep{timem2026} consolidates information across temporal levels, and Mnemis~\citep{mnemis2026} combines similarity retrieval with top-down selection over hierarchical graphs.
Zep~\citep{zep2025} maintains a temporal knowledge graph.
MemoryOS~\citep{memoryos2025} moves memory through short-, mid-, and long-term tiers, and MemOS~\citep{memos2025} manages memory as an operating-system resource.
Infini Memory~\citep{infinimemory2026} maintains topic documents through buffered writing and consolidation; ByteRover~\citep{byterover2026} uses an LLM-curated context tree with explicit relations and provenance.
\name focuses on the consistency of revision: the model edits semantic content, original interactions remain accessible, and the runtime reconstructs access views after accepting a semantic edit.

\subsection{Retrieval and Agent-Controlled Access}

Lexical and embedding retrieval select evidence according to its similarity to a query.
Iterative approaches interleave retrieval with reasoning: IRCoT~\citep{trivedi2023ircot} retrieves with each reasoning step, ReAct~\citep{yao2023react} alternates reasoning traces with tool actions, and Self-RAG~\citep{asai2024selfrag} decides when to retrieve and critiques the retrieved passages.
Progressive observation also appears beyond text: GUI-Lens~\citep{fu2026guilensenablingprecisegui} lets a general-purpose VLM choose the region and scale of each successive view for GUI grounding.
For agent memory, E-mem~\citep{emem2026} reconstructs evidence within selected episodic contexts before answering, and MIRIX~\citep{mirix2025} assigns six memory types to dedicated manager agents coordinated by a meta manager.
MemGPT~\citep{packer2024memgpt} exposes memory operations to the LLM, LSFS~\citep{mei2025lsfs} studies model-operated file access, and DeltaMem~\citep{deltamem2026} trains a single agent with reinforcement learning to manage persona memory end to end.
These methods establish that memory use can include further operations after the first retrieval result.
LLM-as-Code~\citep{qi2026llmascode} argues that deterministic control flow belongs in program code, with the LLM invoked only for reasoning or generation; \name applies a similar division to memory, leaving semantic organization to the model while code validates edits and derives access views.
\name combines continuing access with agent-organized documents and resolvable source references.

\section{Conclusion}

Through \name, we demonstrate the feasibility of LLM agents organizing their own persistent memory.
The agent uses ordinary Markdown documents to decide how information is summarized, grouped, and revised, while the runtime preserves source evidence and maintains consistent access structures.
Experiments on three memory benchmarks demonstrate the effectiveness of this approach, and LoCoMo component analyses clarify the roles of original evidence and organized Topic documents.
The resulting memory can be reused by different Readers and supports further improvements through additional reasoning effort, with lower token consumption in some settings.
These findings show that LLM agents can both read external memory and shape how it is organized and evolves, providing a reusable knowledge base for subsequent tasks.

\section{Limitations}
\label{sec:limitations}

Full-dataset comparisons on three benchmarks and component ablations on LoCoMo primarily evaluate memory-based question answering over conversation histories; they do not fully cover sustained personalization or task execution during long-term interaction.
Reasoning-effort analyses use fixed subsets, and their trends warrant validation on larger question sets and across repeated runs.
Evaluation relies on LLM judges without systematic human calibration and may be affected by judge biases.

Component ablations restrict access to frozen memories rather than testing how memory would be reconstructed or reorganized after removing a component; each condition is a single run, and the overall effects of Core and individual derived views on LoCoMo fall within sampling uncertainty.
Reasoning-effort settings also differ in implementation across models and providers, so their labels do not represent equivalent compute budgets; cross-model comparisons reflect performance under the respective configurations.

Memory construction, maintenance, and multi-round reading incur additional overhead, whose trade-off with answer quality depends on the model, task, and access strategy.
The runtime checks evidence references and structural consistency but cannot guarantee the semantic correctness or completeness of memory content.
Detecting and correcting omissions or errors during continued updates therefore remains an important direction for further study.
Practical deployment additionally requires access controls, retention policies, and isolated tool execution for the stored conversations.

\bibliography{9Reference}

\clearpage
\appendix
\setcounter{topnumber}{5}
\setcounter{dbltopnumber}{5}
\setcounter{totalnumber}{6}
\renewcommand{\topfraction}{0.95}
\renewcommand{\dbltopfraction}{0.95}
\renewcommand{\textfraction}{0.05}
\renewcommand{\floatpagefraction}{0.8}
\renewcommand{\dblfloatpagefraction}{0.8}
\makeatletter
\setlength{\@fptop}{0pt}\setlength{\@fpsep}{12pt plus 2pt}\setlength{\@fpbot}{0pt plus 1fil}
\setlength{\@dblfptop}{0pt}\setlength{\@dblfpsep}{12pt plus 2pt}\setlength{\@dblfpbot}{0pt plus 1fil}
\makeatother
\raggedbottom
\section{Implementation Details}
\label{app:memory_operations}

\paragraph{Validated memory updates.}
Free-form editing by an LLM can silently break the structure that later queries rely on, for example by dropping an evidence reference or pointing a link to a missing paragraph.
\name therefore stages every edit and installs it only after the runtime checks Source immutability, Topic paragraph structure, the resolution of source references and links, temporal values, and allowed paths; Core additionally passes reference and capacity checks.
Accepted edits are normalized, relative links affected by moves are repaired, and Graph, Timeline, Recent, and the file inventory are rebuilt, while a rejected edit leaves the previous state unchanged.
These checks preserve document structure and resolvable evidence links during reorganization; factual interpretation remains the responsibility of the model.

\paragraph{Incremental organization.}
New interactions are archived as Source before the Writer integrates their information into existing Topic and Core content.
Periodic organization revisits Topic files changed since the preceding pass, allowing the agent to revise grouping and document structure as related information accumulates; an optional final pass covers the complete Topic tree.
Both integration and reorganization use the validation procedure above, so the runtime reconstructs access views from the accepted content.
An optional post-write probe asks a factual question about a new interaction and triggers targeted repair when the memory cannot answer it.

\begin{figure*}[tp]
\centering
\includegraphics[width=\textwidth]{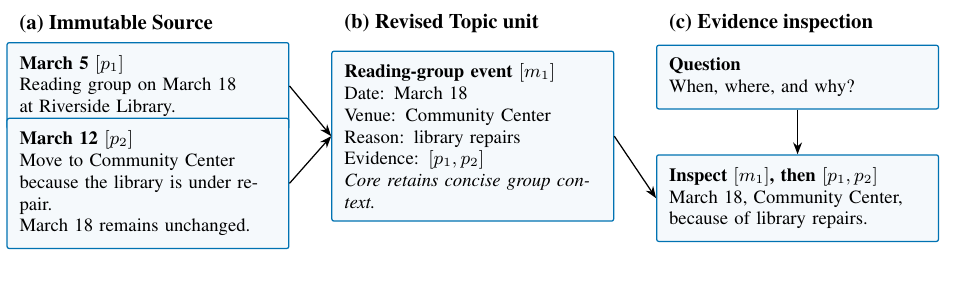}
\caption{An example of a memory update and evidence inspection. A venue change is recorded as a revision of the existing Topic paragraph that cites both the original plan and the relocation notice in Source.}
\label{fig:workspace_example}
\end{figure*}

\paragraph{Query interface.}
At query time, the agent receives Core, Recent, the file inventory, and the question date when available, and can list and read files, run exact-text, BM25, or embedding search, and inspect Source records directly.
No fixed order of operations is imposed, so the agent can enter through a Topic document, a Timeline date, or a search hit, and then follow links and source references as needed.
The split search interface exposes \texttt{bm25\_search} and \texttt{embedding\_search}, together with \texttt{list\_memory\_files} and \texttt{read\_memory\_file}; the latter accepts line offsets and limits.
A combined search interface exposes lexical and dense modes through \texttt{memory\_search}. All main experiments use the split interface.
The read-only shell interface supports file listing (\texttt{ls}, \texttt{find}), literal search (\texttt{rg}, \texttt{grep}), and reading (\texttt{cat}, \texttt{sed}).
These interfaces implement the search, listing, and reading operations described in Section~\ref{subsec:query}.
Figure~\ref{fig:workspace_example} illustrates one update and the corresponding evidence inspection.

\section{Evaluation Details}
\label{app:protocol}

\paragraph{LoCoMo.}
In Table~\ref{tab:main_results}, all memory methods in the primary block use GPT-6 Luna for construction and answering, and all configurations are evaluated on the 1,540 category 1--4 questions with the same GPT-4o-mini judge; \name uses agent-controlled search with at most 20 QA turns.
The primary \name result and its component ablations use GPT-6 Luna as Reader on the same frozen memory. Additional backbone configurations use the named model for both Writer and Reader with their own construction and query settings.

\paragraph{BEAM-100K.}
The seven memory methods in Table~\ref{tab:beam_method_comparison} use GPT-6 Luna for construction and answering on the same 400 questions over 20 conversations.
A-Mem retrieves the top 10 linked notes, and Mem0 and LightMem retrieve the top 20 memories with all-MiniLM-L6-v2 embeddings; MemoryOS, Nemori, and E-mem use their released retrieval pipelines; \name uses agent-controlled search with at most 50 QA turns.

\paragraph{LongMemEval-S.}
All methods in Table~\ref{tab:longmemeval} are evaluated on the complete 500-question set. The primary block uses GPT-6 Luna for construction and answering, while the additional \name row uses DeepSeek V4 Flash for both stages; \name uses agent-controlled search with at most 50 QA turns.

\paragraph{Evaluation protocols.}
All primary evaluations score complete saved answers with GPT-4o-mini at temperature~0.
LoCoMo uses the judging prompt in Figure~\ref{fig:prompt_judge}~\citep{lightmem2026}, with CORRECT/WRONG JSON labels over category 1--4 questions. Overall accuracy is micro-averaged over questions.
LongMemEval-S uses question-type-specific judging prompts: multi-session and single-session-assistant questions use a generic exact-match template, while the remaining types and abstention cases use task-specific templates. The judge uses top-$p=0.95$ and returns a yes/no correctness decision.
BEAM uses rubric-nugget scoring for all capabilities, including event ordering. Each rubric criterion is judged against the question and complete answer. Scores are mapped to 0, 0.5, or 1 using thresholds of 0.25 and 0.75, averaged within each question, and then averaged equally over questions. Pass rates use a question-level threshold of 0.5. The judge response limit is 300 tokens.
Failed requests and invalid LoCoMo labels are unresolved evaluations, not incorrect answers.

\begin{figure*}[tp]
\centering
\begin{promptbox}{LoCoMo Answer Judging}
Your task is to label an answer to a question as `CORRECT' or `WRONG'. You will be given the following data:\\
\quad (1) a question (posed by one user to another user),\\
\quad (2) a `gold' (ground truth) answer,\\
\quad (3) a generated answer\\
which you will score as CORRECT/WRONG.

The point of the question is to ask about something one user should know about the other user based on their prior conversations.
The gold answer will usually be a concise and short answer that includes the referenced topic, for example:\\
\quad Question: Do you remember what I got the last time I went to Hawaii?\\
\quad Gold answer: A shell necklace\\
The generated answer might be much longer, but you should be generous with your grading - as long as it touches on the same topic as the gold answer, it should be counted as CORRECT.

For time related questions, the gold answer will be a specific date, month, year, etc. The generated answer might be much longer or use relative time references (like ``last Tuesday'' or ``next month''), but you should be generous with your grading - as long as it refers to the same date or time period as the gold answer, it should be counted as CORRECT. Even if the format differs (e.g., ``May 7th'' vs ``7 May''), consider it CORRECT if it's the same date.

Now it's time for the real question:\\
Question: \pvar{question}\\
Gold answer: \pvar{gold\_answer}\\
Generated answer: \pvar{generated\_answer}

First, provide a short (one sentence) explanation of your reasoning, then finish with CORRECT or WRONG.
Do NOT include both CORRECT and WRONG in your response, or it will break the evaluation script.

Just return the label CORRECT or WRONG in a json format with the key as ``label''.
\end{promptbox}
\caption{Prompt used for LLM-as-a-Judge evaluation on LoCoMo. The placeholders are filled with the question, the gold answer, and the generated answer of each evaluated item.}
\label{fig:prompt_judge}
\end{figure*}

\section{Supplementary Results}
\label{app:results}

\subsection{LoCoMo Component Ablation Details}
\label{app:full_ablation_results}

Table~\ref{tab:ablation_locomo} reports all four question categories for the component ablations on the complete 1,540-question LoCoMo evaluation set.
The Full configuration uses the same answers and scores as the primary \name row in Table~\ref{tab:main_results}.
The w/o Graph condition removes explicit association links and the Relations index while retaining factual content and Source references.

\paragraph{Controlled settings.}
Table~\ref{tab:experiment_controls} records the frozen memory source and query settings.
All conditions reuse the same memory without reconstruction and change only query-time component access.

\begin{table}[tp]
\centering
\small
\setlength{\tabcolsep}{4pt}
\caption{Controlled settings for the LoCoMo component ablations. The frozen memory, questions, retrieval prompt, search settings, and scoring protocol are shared across conditions.}
\label{tab:experiment_controls}
\begin{tabular*}{\columnwidth}{@{\extracolsep{\fill}}p{0.31\columnwidth}p{0.65\columnwidth}@{}}
\toprule
\textbf{Setting} & \textbf{Value} \\ \midrule
Questions & All 1,540 LoCoMo category 1--4 questions \\
Memory & Frozen memory shared across conditions \\
Reader & GPT-6 Luna, no reasoning effort \\
Query turn limit & 20 \\
Search interface & Split BM25 and embedding search \\
Variable & Query-time component access \\
Judge & GPT-4o-mini, temperature 0 \\
\bottomrule
\end{tabular*}
\end{table}

\paragraph{Context and access.}
Table~\ref{tab:ablation_resources} reports the input to the final answer-producing request and the total number of Reader calls per question.
Removing Recent reduces final input length while leaving average call counts close to Full, supporting the compact-input interpretation in the main text.

\begin{table}[tp]
\centering
\small
\setlength{\tabcolsep}{4pt}
\caption{Mean final-answer input and Reader calls for the LoCoMo component ablations on all 1,540 questions.}
\label{tab:ablation_resources}
\begin{tabular*}{\columnwidth}{@{\extracolsep{\fill}}lrr@{}}
\toprule
\textbf{Condition} & \textbf{Input (K)} & \textbf{Calls} \\ \midrule
Full & 18.57 & 4.85 \\
w/o Topic & 4.67 & 4.31 \\
w/o Core & 17.33 & 4.81 \\
w/o Source & 17.01 & 3.91 \\
w/o Recent & 8.28 & 4.77 \\
w/o Timeline & 18.00 & 4.92 \\
w/o Graph & 18.56 & 4.80 \\
\bottomrule
\end{tabular*}
\par\smallskip\parbox{\columnwidth}{\footnotesize Input is the token count of the final answer-producing request, including instructions, tool definitions, and accumulated dialogue; K denotes 1,000 tokens. Calls is the number of Reader requests over the entire question. Input measures final-request length, not cumulative usage.}
\end{table}

\paragraph{Paired question-level changes.}
Table~\ref{tab:ablation_paired} compares each condition with its matched Full control using the original saved answers and scores on identical questions.
The paired counts distinguish net changes from opposing question-level changes: equal overall accuracy can conceal different sets of correct answers.
We compute paired percentile bootstrap intervals by resampling conversations to preserve dependence among questions from the same history.
These intervals quantify sampling uncertainty conditional on the saved runs.

\begin{table}[tp]
\centering
\small
\setlength{\tabcolsep}{4pt}
\caption{Paired accuracy changes relative to Full on all 1,540 LoCoMo questions. Positive differences favor the ablated condition.}
\label{tab:ablation_paired}
\begin{tabular*}{\columnwidth}{@{\extracolsep{\fill}}lrrrr@{}}
\toprule
\textbf{Condition} & \textbf{$\Delta$ (pp)} & \textbf{W$\to$R} & \textbf{R$\to$W} & \textbf{95\% CI (pp)} \\ \midrule
w/o Topic & $-2.92$ & 65 & 110 & $[-4.96, -0.88]$ \\
w/o Core & $+0.00$ & 52 & 52 & $[-1.16, +1.09]$ \\
w/o Source & $-8.64$ & 28 & 161 & $[-11.67, -6.11]$ \\
w/o Recent & $-1.30$ & 44 & 64 & $[-2.98, +0.30]$ \\
w/o Timeline & $-0.97$ & 48 & 63 & $[-2.54, +0.52]$ \\
w/o Graph & $-0.91$ & 50 & 64 & $[-2.65, +0.85]$ \\
\bottomrule
\end{tabular*}
\par\smallskip\parbox{\columnwidth}{\footnotesize W$\to$R and R$\to$W count wrong-to-right and right-to-wrong changes. Intervals use 10,000 paired percentile bootstrap samples (seed 20261031), resampling the ten conversations. They describe sampling uncertainty for the saved runs, not repeated-run variability; no multiple-comparison correction is applied.}
\end{table}

\subsection{Memory Access Comparison}

\begin{table}[t]
\centering
\small
\setlength{\tabcolsep}{4pt}
\caption{Comparison of memory access methods on LoCoMo with a shared GPT-5.6 Luna Reader. We report LJ accuracy (\%) on the analysis subset. $^{\dagger}$ denotes our local adaptation, and the best results are highlighted in \textbf{bold}.}
\label{tab:locomo_access_diagnostic}
\begin{tabular*}{\columnwidth}{@{\extracolsep{\fill}}lccccc@{}}
\toprule
\textbf{Method} & \textbf{SH} & \textbf{MH} & \textbf{OD} & \textbf{TR} & \textbf{Overall} \\
\midrule
Full Context & 90.91 & \textbf{83.33} & \textbf{83.33} & 90.48 & 89.00 \\
BM25 & 76.36 & 50.00 & 16.67 & 57.14 & 64.00 \\
Dense & 54.55 & 33.33 & 50.00 & 61.90 & 52.00 \\
E-mem$^{\dagger}$ & 85.45 & 72.22 & 50.00 & 85.71 & 81.00 \\
LightMem$^{\dagger}$ & 80.00 & 61.11 & 50.00 & 80.95 & 75.00 \\
\midrule
\name & \textbf{98.18} & 77.78 & \textbf{83.33} & \textbf{100.00} & \textbf{94.00} \\
\bottomrule
\end{tabular*}
\end{table}

\paragraph{Settings and paired outcomes.}
This comparison uses 100 questions stratified over ten LoCoMo conversations and four question types, with a shared GPT-5.6 Luna Reader.
Full Context supplies the original conversation; BM25 and Dense retrieve Source evidence once; E-mem reconstructs routed episodic evidence; and LightMem retrieves extracted facts. The masking conditions restrict access to the same frozen \name memory.
Table~\ref{tab:locomo_access_diagnostic} compares answer accuracy, and Table~\ref{tab:paired_core} reports paired outcomes on identical questions.
The paired results show that organized, iterative access recovers answers missed by one-shot lexical and dense retrieval. Removing Source also loses answers that organized summaries alone do not recover, supporting the use of original records alongside Topic documents.

\begin{table*}[tp]
\centering
\small
\setlength{\tabcolsep}{4pt}
\caption{Question-level paired comparison on the LoCoMo analysis subset between \name and each access method (top) or masking condition (bottom). We report the number of questions in each outcome and the comparator-minus-\name accuracy difference in percentage points with its 95\% bootstrap confidence interval.}
\label{tab:paired_core}
\begin{tabular*}{\textwidth}{@{\extracolsep{\fill}}lrrrrrr@{}}
\toprule
\textbf{Comparator} & \textbf{Both correct} & \textbf{Only comparator} & \textbf{Only \name} & \textbf{Both wrong} & \textbf{Difference} & \textbf{95\% CI} \\
\midrule
Full Context & 88 & 1 & 6 & 5 & $-5$ & $[-10,0]$ \\
BM25 & 64 & 0 & 30 & 6 & $-30$ & $[-40,-20]$ \\
Dense & 52 & 0 & 42 & 6 & $-42$ & $[-54,-32]$ \\
E-mem (adapted) & 81 & 0 & 13 & 6 & $-13$ & $[-19,-8]$ \\
LightMem (adapted) & 74 & 1 & 20 & 5 & $-19$ & $[-31,-9]$ \\
\midrule
w/o Core & 92 & 1 & 2 & 5 & $-1$ & $[-4,2]$ \\
w/o Source & 80 & 2 & 14 & 4 & $-12$ & $[-18,-5]$ \\
\bottomrule
\end{tabular*}
\end{table*}

\subsection{Additional BEAM Results}

\begin{table*}[t]
\centering
\small
\setlength{\tabcolsep}{4pt}
\caption{Pass rate (\%) on BEAM-100K across memory capabilities.}
\label{tab:beam_method_pass}
\begin{tabular*}{\textwidth}{@{\extracolsep{\fill}}lccccccccccc@{}}
\toprule
\textbf{Method} & \textbf{IE} & \textbf{MR} & \textbf{TR} & \textbf{KU} & \textbf{CR} & \textbf{EO} & \textbf{IF} & \textbf{PF} & \textbf{SUM} & \textbf{ABS} & \textbf{Overall} \\
\midrule
Full Context & \textbf{87.50} & 52.50 & 62.50 & 50.00 & 20.00 & 37.50 & 65.00 & \underline{92.50} & \underline{52.50} & \underline{32.50} & 55.25 \\
\midrule
A-Mem & \underline{85.00} & \underline{62.50} & \underline{65.00} & \underline{52.50} & 12.50 & 42.50 & 72.50 & \textbf{95.00} & 40.00 & 30.00 & \underline{55.75} \\
Mem0 & 72.50 & 55.00 & 47.50 & \textbf{62.50} & 22.50 & \underline{50.00} & \underline{75.00} & \textbf{95.00} & 27.50 & 25.00 & 53.25 \\
MemoryOS & 65.00 & \textbf{65.00} & 50.00 & 42.50 & 17.50 & 30.00 & 72.50 & 90.00 & 32.50 & \textbf{42.50} & 50.75 \\
Nemori & 67.50 & 57.50 & 37.50 & 47.50 & 15.00 & 45.00 & \textbf{80.00} & \underline{92.50} & 45.00 & 27.50 & 51.50 \\
LightMem & 65.00 & 57.50 & 47.50 & 45.00 & 27.50 & 25.00 & 72.50 & 85.00 & 15.00 & 30.00 & 47.00 \\
E-mem & 60.00 & 50.00 & 42.50 & 32.50 & \textbf{67.50} & \textbf{52.50} & 62.50 & 90.00 & 15.00 & 17.50 & 49.00 \\
\textbf{\name} & 80.00 & 57.50 & \textbf{77.50} & \underline{52.50} & \underline{40.00} & 47.50 & \textbf{80.00} & \textbf{95.00} & \textbf{60.00} & 25.00 & \textbf{61.50} \\ \bottomrule
\end{tabular*}
\end{table*}

\paragraph{Pass rates by capability.}
Table~\ref{tab:beam_method_pass} complements the nugget scores in Table~\ref{tab:beam_method_comparison} with per-capability pass rates for all methods; \name has the highest overall pass rate and the highest or tied-highest pass rate on temporal reasoning, instruction following, preference following, and summarization.

\par\medskip\noindent
\begin{minipage}{\columnwidth}
\makeatletter\def\@captype{table}\makeatother
\centering
\small
\caption{Results of \name on BEAM-1M with GPT-6.1 Sol as the Reader under two judge models. We report pass rate and mean nugget score (\%).}
\label{tab:beam1m_supplement}
\begin{tabular}{lcc}
\toprule
\textbf{Judge} & \textbf{Pass} & \textbf{Nugget} \\
\midrule
GPT-4o-mini & 69.29 & 63.24 \\
GPT-6 Sol & 84.14 & 73.86 \\
\bottomrule
\end{tabular}
\end{minipage}
\par\medskip

\paragraph{Longer histories on BEAM-1M.}
On BEAM-1M, whose histories are roughly ten times longer than BEAM-100K, \name with GPT-6.1 Sol reaches a 69.29\% pass rate under the GPT-4o-mini judge and 84.14\% under the GPT-6 Sol judge (Table~\ref{tab:beam1m_supplement}).
This experiment supplements the main benchmark with longer histories. Its Reader configuration differs from the primary BEAM-100K comparison, so it documents performance at this scale rather than isolating the effect of history length.

\section{Scaling and Resource Usage}
\label{app:efficiency}

\subsection{Reader Scaling Details}
Tables~\ref{tab:reasoning_details} and~\ref{tab:beam_reader_compute} give the numerical values plotted in Figure~\ref{fig:reader_compute}, on the same 100 LoCoMo and 200 BEAM-100K questions.
Input and output are summed over the Reader calls contributing to each adopted answer and then averaged over questions. Input includes cached, cache-creation, and repeated context; reported reasoning tokens are included in output once. Construction, judging, abandoned attempts, and separate auxiliary session requests are excluded.
These are token-usage measurements, not monetary charges or the length of the final prompt.

Within each sweep, memory and questions are reused. Effort labels denote the requested reasoning settings.

\begin{table*}[tp]\centering\small
\setlength{\tabcolsep}{4pt}
\caption{LoCoMo scaling details corresponding to Figure~\ref{fig:reader_compute}. Score is LJ accuracy ($\times100$). Input, output, and total tokens are means per question in thousands.}
\label{tab:reasoning_details}
\begin{tabular*}{\textwidth}{@{\extracolsep{\fill}}llrrrr@{}}\toprule
\textbf{Reader} & \textbf{Effort} & \textbf{Score} & \textbf{Input (K/Q)} & \textbf{Output (K/Q)} & \textbf{Total (K/Q)} \\ \midrule
GPT-6 Luna & None & 93.00 & 76.609 & 0.210 & 76.819 \\
GPT-6 Luna & Low & 91.00 & 64.998 & 0.184 & 65.182 \\
GPT-6 Luna & Medium & 91.00 & 68.924 & 0.195 & 69.118 \\
GPT-6 Luna & High & 93.00 & 79.253 & 0.247 & 79.500 \\
GPT-6 Luna & xHigh & 91.00 & 95.935 & 0.470 & 96.405 \\
\midrule
GPT-6.1 Sol & Low & 93.00 & 55.897 & 0.155 & 56.052 \\
GPT-6.1 Sol & Medium & 92.00 & 62.888 & 0.170 & 63.058 \\
GPT-6.1 Sol & High & 93.00 & 75.701 & 0.217 & 75.918 \\
GPT-6.1 Sol & xHigh & 95.00 & 98.204 & 0.424 & 98.628 \\
\midrule
Qwen 3.8 Flash & Low & 92.00 & 54.760 & 0.384 & 55.144 \\
Qwen 3.8 Flash & Medium & 94.00 & 71.242 & 0.539 & 71.781 \\
Qwen 3.8 Flash & xHigh & 94.00 & 84.926 & 0.699 & 85.626 \\
\midrule
GLM-5.3 Flash & Low & 93.00 & 48.503 & 0.236 & 48.739 \\
GLM-5.3 Flash & High & 94.00 & 50.010 & 0.303 & 50.313 \\
GLM-5.3 Flash & Max & 94.00 & 60.480 & 0.623 & 61.103 \\
\bottomrule\end{tabular*}\end{table*}

\begin{table*}[tp]\centering\small
\setlength{\tabcolsep}{4pt}
\caption{BEAM-100K scaling details corresponding to Figure~\ref{fig:reader_compute}. Score is mean nugget score ($\times100$). Input, output, and total tokens are means per question in thousands.}
\label{tab:beam_reader_compute}
\begin{tabular*}{\textwidth}{@{\extracolsep{\fill}}llrrrr@{}}\toprule
\textbf{Reader} & \textbf{Effort} & \textbf{Score} & \textbf{Input (K/Q)} & \textbf{Output (K/Q)} & \textbf{Total (K/Q)} \\ \midrule
GPT-6 Luna & None & 57.85 & 115.989 & 0.557 & 116.547 \\
GPT-6 Luna & Low & 56.82 & 47.621 & 0.468 & 48.089 \\
GPT-6 Luna & Medium & 57.60 & 75.804 & 1.015 & 76.820 \\
GPT-6 Luna & High & 59.07 & 110.884 & 1.675 & 112.558 \\
GPT-6 Luna & xHigh & 61.79 & 102.774 & 1.637 & 104.411 \\
\midrule
GPT-6.1 Sol & Low & 63.98 & 38.223 & 0.490 & 38.713 \\
GPT-6.1 Sol & Medium & 66.35 & 44.427 & 0.572 & 45.000 \\
GPT-6.1 Sol & High & 67.25 & 63.551 & 0.818 & 64.369 \\
GPT-6.1 Sol & xHigh & 70.42 & 101.481 & 1.849 & 103.330 \\
\midrule
Qwen 3.8 Flash & None & 70.83 & 202.039 & 2.307 & 204.346 \\
Qwen 3.8 Flash & Low & 66.68 & 194.662 & 2.845 & 197.507 \\
Qwen 3.8 Flash & Medium & 69.60 & 157.804 & 2.429 & 160.233 \\
Qwen 3.8 Flash & xHigh & 70.35 & 207.606 & 2.684 & 210.290 \\
\midrule
GLM-5.3 Flash & Low & 63.93 & 20.188 & 0.527 & 20.715 \\
GLM-5.3 Flash & High & 68.82 & 25.975 & 0.791 & 26.766 \\
GLM-5.3 Flash & Max & 70.43 & 46.652 & 1.999 & 48.651 \\
\bottomrule\end{tabular*}\end{table*}

\subsection{Construction and Query Resources}
The following resource measurements supplement the Reader-only scaling analysis. Construction is counted once, whereas query resources accumulate over answered questions.

\begin{figure*}[tp]
\centering
\includegraphics[width=\textwidth]{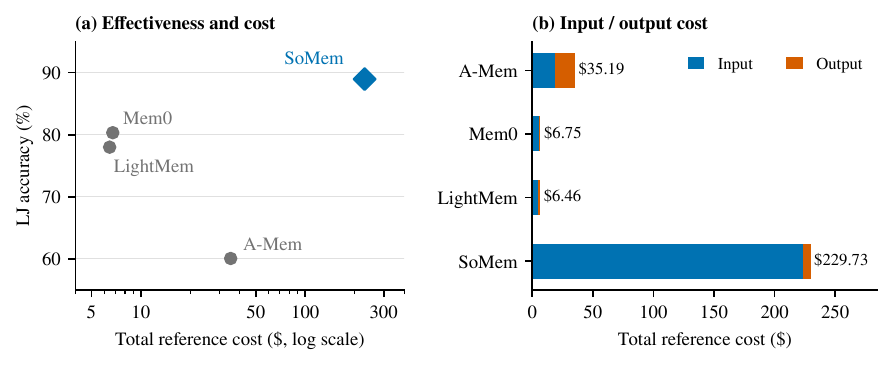}
\caption{Accuracy and cost of memory methods on the full LoCoMo benchmark with GPT-6 Luna. (a) LJ accuracy versus total cost on a logarithmic axis. (b) Cost contributions of input and output tokens, priced at \$1/\$6 per million.}
\label{fig:cost_tradeoffs}
\end{figure*}

\begin{table}[tp]
\centering
\small
\setlength{\tabcolsep}{4pt}
\caption{Query cost of two Readers on the full LoCoMo benchmark: LJ accuracy, model calls per question (Calls/Q), and mean end-to-end latency per question.}
\label{tab:query_resources}
\begin{tabular*}{\columnwidth}{@{\extracolsep{\fill}}lrrr@{}}
\toprule
\textbf{Reader} & \textbf{Acc. (\%)} & \textbf{Calls/Q} & \textbf{Latency (s)} \\
\midrule
GPT-5.6 Luna & 92.34 & 5.21 & 21.59 \\
DeepSeek V4 Flash & 93.18 & 7.40 & 18.79 \\
\bottomrule
\end{tabular*}
\end{table}

\paragraph{Cost is concentrated in reusable input context.}
On the full LoCoMo benchmark, \name reaches 88.96\% LJ accuracy at a total reference cost of \$229.73, compared with \$6.75 for Mem0 (80.32\%), \$6.46 for LightMem (77.99\%), and \$35.19 for A-Mem (60.06\%) (Figure~\ref{fig:cost_tradeoffs}).
Of this cost, \$118.85 is one-time construction, and querying costs about \$0.07 per question.
Input context dominates the usage, and 66.3\% of input tokens are cache hits across the recorded calls; charging cached input at 10\% of the input price lowers the total to \$96.33.
The cross-model reuse experiment in Section~\ref{sec:exp:cross_model} shows how a construction cost can be shared across subsequent queries and Readers.
In separate Reader runs, querying averages 5.2--7.4 model calls and about 20 seconds per question (Table~\ref{tab:query_resources}).

\end{document}